# Toward Data Systems That Are Business Semantic Centric and AI Agents Assisted

**Cecil Pang**

School of Systems Science and Industrial Engineering, Binghamton University, State University of New York, Binghamton, NY, USA

e-mail: cpang2@binghamton.edu

**ABSTRACT** Contemporary businesses operate in dynamic environments requiring rapid adaptation to achieve goals and maintain competitiveness. Existing data platforms often fall short by emphasizing tools over alignment with business needs, resulting in inefficiencies and delays. To address this gap, I propose the Business Semantics Centric, AI Agents Assisted Data System (BSDS), a holistic system that integrates architecture, workflows, and team organization to ensure data systems are tailored to business priorities rather than dictated by technical constraints. BSDS redefines data systems as dynamic enablers of business success, transforming them from passive tools into active drivers of organizational growth. BSDS has a modular architecture that comprises curated data linked to business entities, a knowledge base for context-aware AI agents, and efficient data pipelines. AI agents play a pivotal role in assisting with data access and system management, reducing human effort, and improving scalability. Complementing this architecture, BSDS incorporates workflows optimized for both exploratory data analysis and production requirements, balancing speed of delivery with quality assurance. A key innovation of BSDS is its incorporation of the human factor. By aligning data team expertise with business semantics, BSDS bridges the gap between technical capabilities and business needs. Validated through real-world implementation, BSDS accelerates time-to-market for data-driven initiatives, enhances cross-functional collaboration, and provides a scalable blueprint for businesses of all sizes. Future research can build on BSDS to explore optimization strategies using complex systems and adaptive network theories, as well as developing autonomous data systems leveraging AI agents.

**INDEX TERMS** data system, data architecture, data engineering, AI agent, business semantics, data science, machine learning

## I. INTRODUCTION

### A. BACKGROUND

In the current business landscape, in order to catch growth opportunities or merely stay competitive in the ever-changing markets, businesses need to formulate strategies, implement new tactics, and optimize operations in a timely fashion. This is an iterative process of ideation, experimentation, decision-making, and implementation. Furthermore, once a new tactic has been rolled out, it must be monitored to ensure the expected level of performance. The above iterative process needs to be repeated, in its entirety or partially, when the metrics indicate deviation from the expectation. This process involves multiple business units, and the end-to-end time window available is often very short—days rather than weeks or months. For modern businesses, this process is typically data-driven and involves analytics and data science modeling rather than being solely based on intuition and domain expertise. In order to enable such a data-driven business at scale, a flexible and efficient data system is a must. For example, imagine a digital content subscription business aiming to improve the retention of subscribers by offering promotional deals. It is an iterative, data-driven process that requires the collaboration of multiple departments: the retention team define the criteria for the target audience as well as the promotions to offer, the data scientists build models to identify the target audience, the marketing technologies team send the promotions via marketing platforms, and data engineers make data available to support all of the above steps. The availability of necessary data promptly is often the critical factor that determines the success or failure of such a business objective.

Giebler et al. (2021) [1] stated that defining a generic data architecture is difficult because there were too many possible kinds of data and how they were used. Because of this complexity, it requires businesses to have big upfront investments in data management capabilities as well as a long wait time before the impact of data can be realized. In addition, the complexity often makes data platforms difficult to use, especially for non-technical users, and therefore limits their effectiveness. It is also a common cause of friction among different teams, such as data engineering, data science, and business units, which further reduces their effectiveness. The complexity of a modern data platform arises not only from the inherent intricacies of the data itself but also from the challenges of making that data accessible and actionable. "We need to postpone this strategic project for a quarter because the data is not yet available…" and "We have no choice but to shift our resources to a less impactful project while we are waiting for the data…" are common complaints that have become the norm in many companies. The availability of data determines when business needs can be addressed. In other words, business priorities are determined by data availability. It should be the other way around.

### B. THE PROPOSED SYSTEM

I present a business semantics-centric, AI-assisted data system ("BSDS"). In order to enable today's data-driven business, a data system must center around what a business does to achieve its business goals. As Goodhue et al. (1988) [2] pointed out, the ultimate goal of data resource management is to enable the needs of the business rather than just putting tools in place. I further suggest that having good tools only is not enough. BSDS is a system that consists of tools and technologies, workflows, and people. It is an integral part of a data-driven business rather than only a data management tool.

BSDS is designed to prioritize business needs and objectives, accelerating time to market. With its low barrier to entry, BSDS is accessible to organizations of all sizes. Moreover, it offers high scalability, enabling businesses to rapidly achieve impactful results while providing the flexibility to iteratively adapt and address more complex requirements as they evolve. I take a holistic approach, and BSDS consists of three fundamental and complementary components:

1) A business semantics-centric, AI agents-assisted data architecture (section II),
2) Business goals-focused workflows to effectively utilize the data architecture (section III),
3) Data team organization to optimize system effectiveness and productivity (section IV)

I validated the effectiveness of BSDS by applying it in a real business setting, and the empirical result is provided in section VI as a case study.

### C. RELATED WORK

There is a substantial body of literature on big data management, such as [3], [4], [5], [6], and its support of data science and machine learning [7], [8], [9], [10]. I also find in the literature the concept of a data lake and its architecture and management, such as [11], [12], [13]. However, these works focused primarily on introducing novel capabilities. Little attention was given to helping companies navigate the complexities to achieve business goals or the people factor. Both of these are critical to the success of a data platform.

There is literature on semantics and conceptual modeling in database design. For example, Storey (1993) [14] used entity relationships (E-R) modeling to capture the semantics of the real world and incorporate that into the data stored in the database. Sukhobokov et al. (2022) [11] stated that the structure of the data stored should be semantically ordered. Embley and Liddle (2013) [15] proposed that conceptual modeling might help address some of the volume, variety, velocity, and veracity challenges of big data. Jarke and Quix (2017) [16] traced the evolving role of conceptual modeling since the late 1970s and proposed a vision of interacting data spaces.

Research on entity relationship modeling itself can be dated back to a research paper published in 1976 [17]. Since then, entity relationship modeling has been the foundation of relational database design for decades [18].

Regarding AI agents, agentic systems powered by large language models (LLM) have been making tremendous progress recently [19], [20]. Not only are LLMs becoming more capable of automating tasks and being context-aware, agentic system frameworks that enable the development of multi-agent systems (MAS) are becoming available [21], [22], [23], [24]. With that, LLM-based MASs are being developed in various domains. Some examples are multi-agent robot task planning systems using LLMs [25], modeling interactions among MAS agents using strategy recommendations from LLM [26], and MAS for software engineering [27], [28]. While LLM-based MAS is very new, research on autonomous agents itself can be dated back to the early 1990s. Maes (1993) [29] described autonomous agents as a system operating in a dynamic environment to achieve time-dependent goals, and adaptive autonomous agents improve competence over time. Franklin and Graesser (1997) [30] proposed a formal definition of an autonomous agent, which distinguished it from being just a software program. Since then, much research has been done, and there are textbooks on MAS [31], [32], [33].

Few papers have offered clear guidance for applying a modern data system to solve actual business problems, which should be the ultimate goal of data resource management [2]. BSDS's holistic approach enables a data-driven business with data architecture, workflows, and data team organization.

### D. CONTRIBUTION

The contribution of this work is twofold:

1) Practical Implementation: I provide clear and detailed descriptions of all the components of BSDS as well as design principles that businesses can easily follow, making it practical for immediate use.
2) Holistic System Perspective: By offering a comprehensive system view encompassing people,

workflows, and architecture as well as clearly defined components and interactions, I lay the groundwork for further research in planning and optimization of data systems through complex systems and adaptive network theories [34], while also paving the way for the research of autonomous data systems.

## II. ARCHITECTURE

I propose a modular architecture, as shown in Figure 1. Data consumers are assisted by AI agents to access the curated data. The AI agents are backed by a knowledge base of data and business semantics. Data pipelines load data from multiple sources and perform various levels of transformations and integrations to prepare them to be consumed. The following guiding principles are adhered to during the design of these modules:

1) Simplicity: A simple system is more likely to be useful and adopted.

2) Business Entities-Centric: Ensures alignment with business needs by focusing on core business entities rather than simply creating sophisticated tools.

3) Facilitate AI Agent Integration: Automate system processes to reduce human efforts, thereby accelerating implementation of business use cases, reducing time-to-market, and improving system scalability and operational efficiency.

4) Low Barrier to Entry with Incremental Scalability: Designed to be accessible for businesses of all sizes, ensuring ease of adoption while allowing for seamless growth.

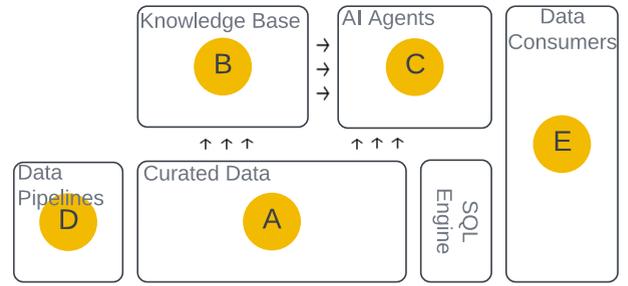

**FIGURE 1.** Modules of BSDS: Data consumers are assisted by AI agents to access the curated data
  A.  **Curated Data:** cleaned, linked, and ready to be consumed
  B.  **Knowledge Base:** metadata, semantics, and people
  C.  **AI Agents:** assist in accessing and managing data, using the knowledge base as context
  D.  **Data Pipelines:** curate data by performing extract, load and transform (ELT) on data from multiple sources
  E.  **Data Consumers:** people, BI tools, and data science models

Detailed compositions of the modules are shown in Figure 2 and are described in detail in the following sub-sections.

### A. CURATED DATA

At the core of curated data is master data. Master data define the key business entities that a company's operations rely on and are utilized throughout the organization [35]. Some typical examples include customers, products, and suppliers. Based on the guiding principles, curated data should be business entities-centric, which means that they should be linked to the master data by way of foreign keys to the business entities. Master data is normalized [18] to facilitate linking as well as lifecycle management [35].

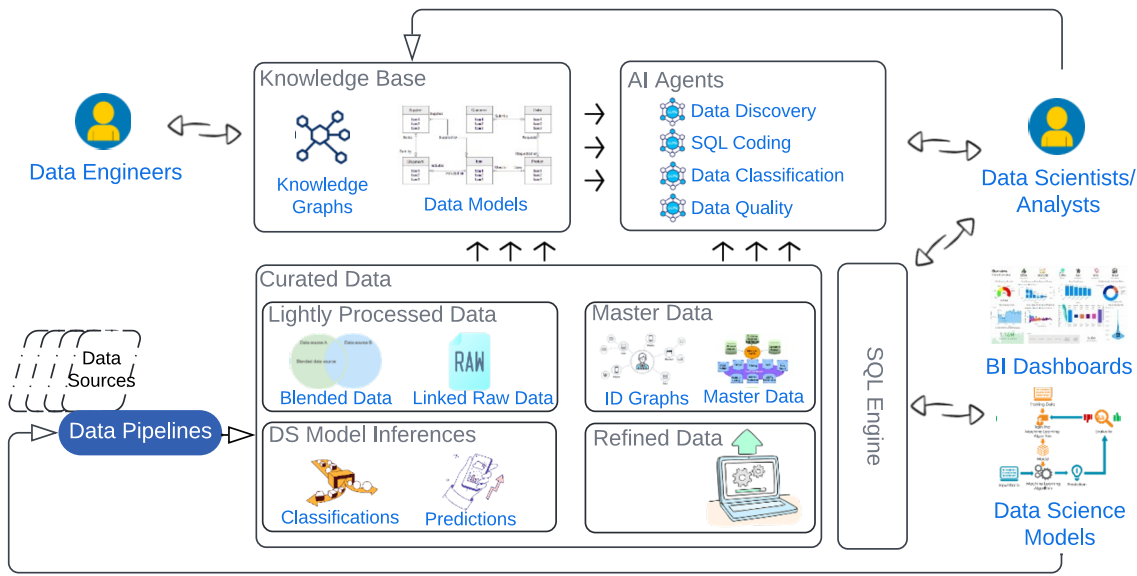

**FIGURE 2.** An overview of the Business Semantic-Centric Data System (BSDS) architecture, illustrating its key components: curated data centers around normalized master data and ID graphs; a knowledge base of data models and knowledge graphs of business entities, datasets and people; AI agents for assisting data consumption and management. Together, these components enable scalable, automated, and context-aware data utilization across the organization.

To further facilitate linking to master data, I propose creating ID graphs for business entities that are identified by different keys in multiple datasets. For example, a consumer may be identified by a user ID in the web store and at the same time the billing system assigns an account number to the same person. In this scenario, a unique identifier called consumer ID is assigned to each consumer, and it is mapped to the web store user ID and the billing system account number in the ID graph. The consumer ID will then be used as the foreign key to link to the consumer entity in the master data. The entity-relationship diagram of such an ID graph is illustrated in Figure 3.

With the master data and ID graphs in place, various datasets are linked to the business entities by adding foreign keys to the right places. I call this lightly processed data. Another common light processing is the blending of closely related data from multiple sources. Using an online content subscription business as an example, there may be more than one business system where subscriptions are captured due to an acquisition or merger. Blending of the data to make one single subscriptions dataset may make it easier for the data to be used. However, it is often okay not to blend the data and leave them as separate datasets in favor of making the data available sooner.

Data science model outputs such as predictions and classifications are often useful additions to the curated data. These new data elements are linked back to the business entities and made available to all data consumers. Similarly, there may be certain aggregations, calculations, and metrics alike that are frequently used or that the underlying business logic needs to be consistent across the board. These are good candidates to be added to the system. I called this refined data.

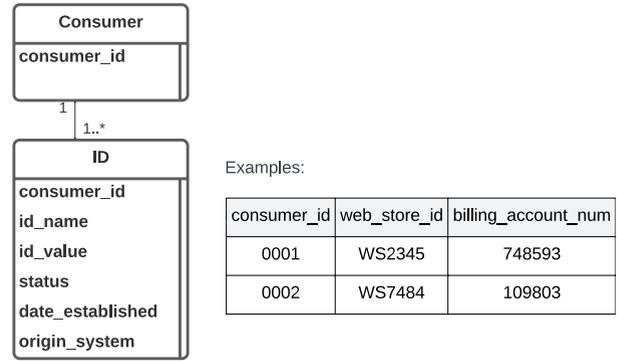

**FIGURE 3.** An entity-relationship diagram illustrating the use of ID graphs to link multiple identifiers (e.g., web store ID, billing account number) to a unified master data entity (Consumer). This approach enables consistent referencing and integration of curated data across disparate systems by normalizing identifiers through a central consumer ID.

Over time, the system grows with more data linked to the business entities and becomes more useful to enable a wide variety of business use cases. Figure 4 is an illustration of multiple datasets linked to a key entity, the Consumer, for an online content subscription business.

### B. KNOWLEDGE BASE
The data model of the business entities and their linked data serves as a knowledge graph of the curated data [36]. Large Language Model-based autonomous agents (AI agents) [20] can utilize this knowledge graph as context for various data access and management tasks. As datasets are linked to the entities in master data, they will be picked up and utilized by the AI agents automatically. I also suggest creating data models for datasets that are not yet loaded in the system but can be useful in the foreseeable future. This is very helpful in data discovery for planning, prioritization, and gap analysis purposes. Figure 5 is an illustration of data models of potentially available datasets connected to a business entity.

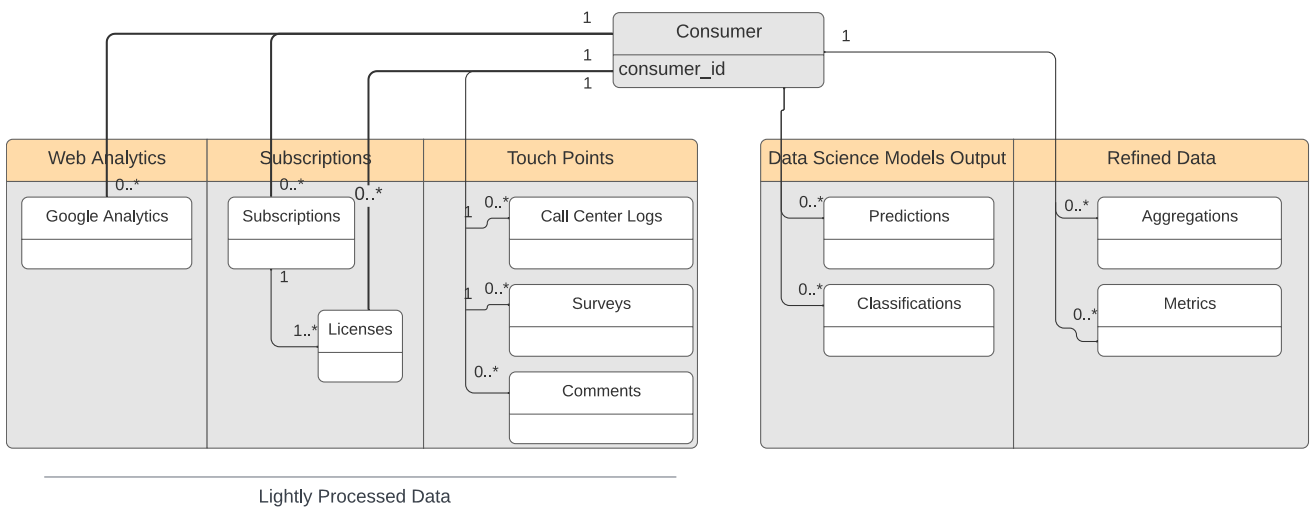

**FIGURE 4.** Curated data organized around the Consumer entity in an online content subscription business. Various datasets—such as web analytics, subscriptions, and touchpoints—are linked via foreign keys using the ID graph that unifies consumer identifiers across systems. Data science model generated predictions and classifications, as well as refined data such as commonly used aggregations and standardized metrics are linked back to the Consumer entity. Over time, the curated data expands in both scope and utility, enabling a wide range of analytical and operational use cases.

Furthermore, I extend the knowledge graph to incorporate the people element, acknowledging its essential contribution to the system's success. This integration facilitates the consideration of human factors in the planning and optimization of BSDS, resulting in a more holistic and adaptive system. I will discuss this more in the Human Factor section.

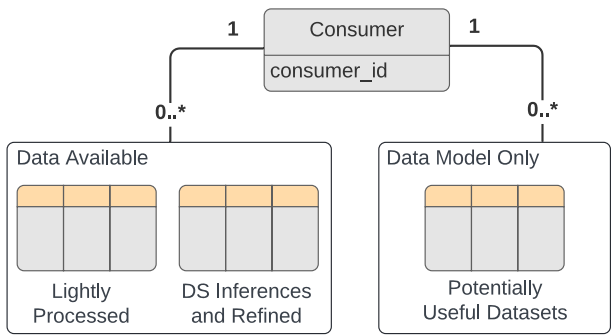

**FIGURE 5.** Illustration of a knowledge base formed by connecting data models of both existing and potential datasets to a central business entity (Consumer). This structure supports data discovery, planning, and gap analysis.

### C. AI AGENTS
In BSDS, AI agents are first-class system components. They translate business intent into executable actions and continuously adapt the platform to new requirements. With a knowledge base serving as the context and the rapid advancement of context-aware LLM agents, there are numerous opportunities for utilizing AI agents in BSDS. Here, I highlight two categories that can easily be included in BSDS: agents to assist data consumers in using the data in the system and agents that help manage the system. For example, an AI agent can assist data scientists and analysts in searching for data immediately available, as well as data that can be made available on request. Another example of assistance that an AI agent can provide to data consumers is generating SQLs to access the curated data. This can be a productivity booster, especially during data exploration.

For the management of data in the system, there are AI agents that classify and label data based on the data model and sampling of the actual data. Classification of data is important for compliance, risk management, and many other purposes. AI agents can also assist in data quality management by generating data quality rules. It reduces the human effort involved in writing the rules as well as providing more comprehensive coverage by automatically detecting new datasets.

Figure 6 presents the multi-agent system architecture of BSDS that leverages LLM agents to automate and validate data access and management tasks in a business semantic-centric environment, with human-in-the-loop escalation. The process is initiated when a user submits an instruction in natural language, or by a scheduled predefined instruction. This instruction, together with relevant context from the knowledge graph and previous outcomes, enters the system and is processed by the following sequence of agents:

- **Generation Agent:** The generation agent interprets the instruction and produces a concrete, machine-executable output (e.g., a SQL statement or an API call with specific arguments), along with a step-by-step reasoning trail that explains how the output was derived.
- **Routing Agent:** The routing agent receives the generated result and reasoning, and dispatches them to a pool of verification agents for independent evaluation.
- **Verification Agents:** Multiple verification agents independently analyze the generated result and reasoning, leveraging the knowledge graph to access semantic correctness and alignment with the data model. Each agent provides a confidence score along with its assessment. The verification agents are typically implemented using different LLMs than the generation agent, and often from different providers.
- **Decision and Escalation:** If the collective confidence from the verification agents exceeds a predetermined threshold, the result is accepted as valid. If confidence is low or if any agent flags concerns, the system escalates the output for human expert validation and potential revision. Human reviewers can also provide feedback or comments, which are captured for system improvement. The routing agent consults the knowledge graph to identify the human experts who have expertise in the business entities and datasets involved in the task.

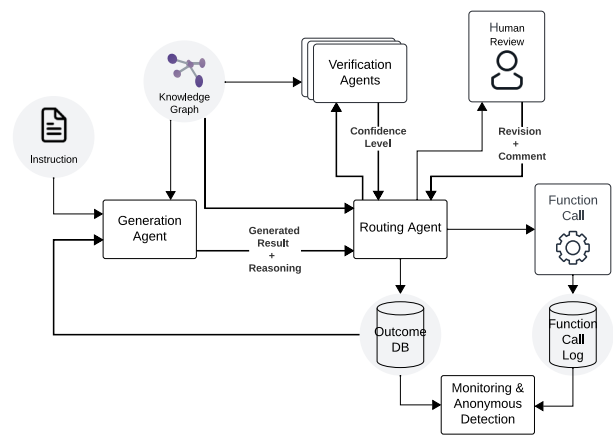

**FIGURE 6.** Architecture of the multi-LLM agent system in BSDS. Instructions are processed by a Generation Agent, evaluated by multiple Verification Agents with confidence scoring and semantic grounding via the knowledge graph, and escalated to human review when needed. All results and decisions are logged for traceability, real-time anonymous detection and continual improvement.

- **Monitoring & Anonymous Detection**: All function calls, agent outputs, verification results, and human feedback are systematically recorded in the function call log and the outcome DB. This provides full traceability, supports monitoring and anonymous detection, and allows for continuous learning.

Throughout this workflow, the knowledge graph provides essential business semantics and data relationships, grounding all agent reasoning in accurate business context. In addition, the outcome DB and function call log provide insights on how work actually happened, guiding subsequent agent actions. This multi-agent system design ensures that automation is reliable, explainable, and business semantic aligned, while maintaining strong safeguards through validation agents' consensus and human oversight when necessary. In addition, the outcome database and function call log provide data for real-time anonymous detection, further mitigating the risks associated with relying on AI agents for critical data access and management tasks. Anonymous detection utilizing multi-sensor time-series data is an active research topic, such as [37] and [38].

BSDS takes a hybrid approach in the real-time selection and orchestration of agents and humans in the loop. Rule-based routing is used for the standard procedure of generation agent, verification agent, human in the loop, and logging. For tasks that require multi-step workflows, a planner agent dynamically determines the steps and route each step to the rule-based standard procedure. The rule-based standard procedure makes the system predictable and easy to audit—a must for data systems—while the planner agent takes care of complex tasks and adapts to new scenarios with minimum manual intervention.

Multi-LLM agent systems can be computationally demanding. Each LLM-powered agent consumes notable resources, and end-to-end workflows can introduce significant latency and cost per request. To mitigate these challenges, BSDS

- employs rule-based standard procedure to avoid unnecessary LLM calls,
- prefers smaller specialized models over large general-purpose models for well-defined tasks such as SQL generation, and
- adheres to distributed system best practices such as asynchronous calls and parallelizing where possible.

### D. DATA PIPELINES
Datasets typically arrive at BSDS from multiple sources. These sources can be from business partners and vendors or internal software applications. While there are many data pipeline tools available in the market, I recommend a simplistic approach of doing as much as possible using SQL. Depending on the source, there may be limited choices of tools to extract and load the data in BSDS. However, once the data have been loaded, the transformations that connect the new data to the entities in the master data should be done using a SQL-based tool. Standardizing in SQL encourages data engineers to focus on the data and business semantics and not be distracted by the tools. It also has the advantage that, in addition to the data models, the SQLs can be used as context to enable AI agents, for example, to generate data lineage.

### E. DATA CONSUMERS
BSDS can support a wide variety of data consumers. People, business intelligence (BI) tools, training and inferencing of data science models, external AI agents, etc. However, regarding tools, I suggest avoiding unnecessary variations and choosing tools that have similar data structure requirements or preferences. This is especially true when the different options are similar in all other important considerations and therefore there is no real benefit of using multiple tools. This approach minimizes the data transformations and refinements required to support all the tools.

### III. OPERATION WORKFLOWS
I propose two workflows, one for data exploration and experimentation and the other for fulfilling concrete data requirements. The primary focus of the former is to make available data for exploration quickly, while the objective of the latter is to prepare data for ongoing use by various types of data consumers in a production setting. This two-workflow approach promotes speedy delivery of many datasets for exploration, not wasting extra effort on data that would not be used in production and, at the same time, maintaining high standards for putting data in production. Traditionally, data teams use the same workflow to fulfill these two very different requirements. The results are either being too slow for the former or not having enough rigor for the latter. Often both.

These two workflows, though very different, are connected in a way that the result of the former will be useful to the latter.

### A. DATA EXPLORATION
An example of the exploratory workflow is when a data scientist tries to explore various data elements that may be correlated with a certain prediction, such as the probability that a consumer would stop the subscription. This kind of usage typically requires access to many data elements in several datasets. At the end of the exploration, usually only a small portion of the data explored would be selected to feed the final data science model. A quick turnaround time to make the data available for exploratory analysis and experimentation is critical to the success of a data-driven business. To make data useful for exploration, the raw data is connected to the entities in master data by ID graphs lookup and then adding foreign keys. Other than that, just enough checking and cleaning is done to ensure that the data is not wrong, with the understanding that the target data consumer may need to write extra lines of SQL to make the data useful for exploratory purposes. Since this workflow trades data quality for speed, it is expected that there may be additional data cleaning and processing needed during and after the exploratory phase. As such, these datasets are labeled as suitable for exploration only.

### B. DATA FOR CONCRETE REQUIREMENTS

This workflow is used towards the completion of data exploration when it becomes clear what data are needed, or to fulfill a well-defined data requirement from the very beginning. Unlike the data exploration workflow, the primary objective of this workflow is to ensure that the data in production is of high quality and is suitable to be consumed by multiple types of tools, such as business intelligence (BI) dashboard builders.

To achieve high quality, more rigorous data cleaning and quality assurance based on data engineering best practices are performed on the datasets. Moreover, the BI tools may have preference on how the data is structured, and therefore, additional data transformation may be necessary to support these tools. Also included in this workflow is adding refined data, such as common predefined metrics and data aggregations.

## IV. HUMAN FACTOR

### A. DATA TEAM ALIGNMENT

The human factor is integral to the success of BSDS. BSDS strategically aligns the data team with business semantics in order to efficiently manage and orchestrate the workflows. The following guiding principles underpin this approach:

1) **Integrated Responsibility**: The data team (engineers, scientists, and analysts), together with other business units, collectively take responsibility for developing and implementing data-driven business strategies, extending beyond the traditional role of merely delivering technical services.

2) **Semantic Expertise**: A deep understanding of data in relation to business semantics is essential, ensuring that technical solutions are both contextually relevant and aligned with business objectives.

BSDS places business semantics at its core, influencing not only its architecture and workflows but also the organization of the data team. As business priorities evolve, the data team must effectively contribute as integral members of business objective-driven projects and experiments, often with minimal time to acclimate. Therefore, a strong understanding of business semantics is crucial for these roles. In addition to possessing a general knowledge of the business and its industry, I suggest that each member of the data team develop expertise in one or more key business entities and their associated data. This is essential to leaving no gap between technical solutions and business objectives. While proficiency in tools and technologies is considered a fundamental prerequisite, assuming robust technical skills among data professionals, I stress that technical expertise alone is insufficient. Figure 7 is a graph illustrating connections between business entities, datasets, and members of the data team. With the human factor added, the graph can be used to plan and optimize the entire system.

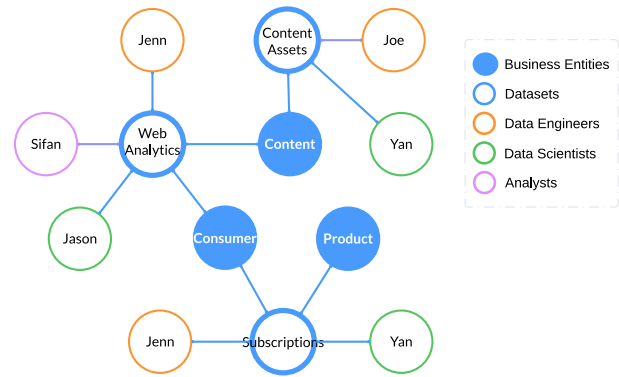

**FIGURE 7.** Illustration of an extended knowledge graph incorporating data team members, business entities, and datasets—highlighting the human factor in aligning technical solutions with business objectives.

### B. EFFECTIVE ALIGNMENT FACTORS

The effectiveness of aligning data team expertise with business semantics can vary significantly across organizations, and this alignment is a critical factor for the success of BSDS. The two common factors that influence how well this alignment is achieved are:

- **Business acumen of the data team**: Many companies overemphasize technical proficiency, often at the expense of business understanding. This can result in having technically skilled teams that lack the ability to interpret data within its business context.

- **Data team culture**: Data teams are often structured around tasks and projects rather than data ownership. This task-oriented culture encourages members to focus only on the data relevant to their immediate responsibilities. As a result, teams often struggle to "connect the dots" and lack a sense of ownership over the data—hindering their ability to derive broader business insights.

### C. ACTIONABLE STRATEGIES

To ensure effective alignment of data team expertise with business semantics, organizations can implement the following actionable strategies:

1) Embed **semantic responsibility into roles**: Use the extended knowledge graph (illustrated in Figure 7) as

a semantic responsibility matrix to assign and track ownership of business entities and their datasets.

2) Reshape **hiring and onboarding practices**: Redesign hiring to assess candidates' business acumen. Incorporate business-context case studies into interviews. Prioritize individuals who can reason with data in a business context. During onboarding, require new hires to complete a rotation or shadowing period with key business units to absorb domain context early.

3) **Foster a culture of data ownership:** Shift from a task-based delivery model to an ownership model in which the data team serves as stewards for key business entities and their related data. Appoint leaders within the team to take responsibility for the data associated with specific business entities, ensuring clear accountability and long-term data quality.

4) **Implement a targeted training strategy:** Develop a structured training program to build semantic fluency across the data team. The program covers:
   a. business literacy—key business entities, functions, terminologies and KPIs
   b. data interpretation—how business events map to data models
   c. case studies—realistic scenarios requiring both business understanding and data modeling. E.g. "Understanding Customer Churn from a Data Perspective"

## V. LOW BARRIER TO ENTRY & SCALABLE

Scalability is embedded across the architecture, workflows, and team structure of BSDS. This section demonstrates how scalability is realized through a low barrier to entry and seamless scalability as business needs expand.

### A. GETTING STARTED WITH BSDS

BSDS has a low barrier to entry and requires low upfront investment by providing a simple path to get started. Consider a startup aiming to leverage data science to fuel its growth. BSDS can be deployed through a simple two-step process:

1) **Assemble a lean data team** of two roles: a data engineer and a data scientist. A team of just two people is sufficient to get started. Since BSDS is designed to be implemented on any major public cloud using common, widely adopted services, the required skill sets are commonly available and easy to hire for. Online training courses are widely available from the cloud providers as well as third party learning platforms at low cost.

2) **Deploy BSDS on a public cloud platform** such as Google Cloud Platform (GCP), Amazon Web Services (AWS), or Microsoft Azure. BSDS is cloud-agnostic and does not mandate a specific provider. At the outset, beyond essential cloud practices like permission management and cost monitoring, the only infrastructure required is a SQL-based database or data warehouse (e.g., BigQuery on GCP). Many startups already utilize public cloud platforms to host their products, further reducing the initial barrier to adoption. It is important to note that AI agents are not required from the outset and can be integrated at any stage. As long as the data team and datasets remain aligned with business semantics, AI agents can be added whenever the organization is ready.

### B. SCALE AS BUSINESS GROWS

Consider two common scenarios of growth:

1) **A surge in data volume**—such as exponential increases in customer acquisitions.

2) **The emergence of more complex use cases**—requiring the integration of numerous new datasets.

BSDS is designed for cloud-native deployment, leveraging commonly adopted cloud services. The first scenario is readily managed by modern cloud data platforms, which are built to scale storage and compute resources efficiently.

The second scenario—integrating a growing number of diverse and evolving datasets—is more complex. BSDS's business semantics-centric architecture logically connects datasets to core business entities via the ID graphs. Physically, each dataset can be stored and scaled independently by the underlying cloud infrastructure. This architecture enables several key advantages:

- New datasets and queries can be added without disrupting existing functionality.
- Existing queries can be extended by joining with new datasets, without altering the original logic.
- Each dataset can scale independently

As such, the overall system can grow without requiring major changes to its core design—leading to overall scalability.

## VI. VALIDATION CASE STUDY

BSDS was implemented at an online content subscription business in the US as a proof of concept. I present it here as a case study of how the traditional approach took too long to enable a retention campaign, the key differences that contributed to BSDS's success, the time-to-market metric evaluation, and finally the generalization of the case study.

### A. INTRODUCTION AND PROBLEM STATEMENT

One of the top priorities of the company was to increase the retention of subscribers and reduce the churn rate. The retention team formulated strategies and designed an experiment to proactively reach out to high-churn-risk subscribers and offer them promotions. Then, they made a request for the data science team to identify the high churn risk subscribers. After several working sessions and consulting a few other subject matter experts, a list of data elements were identified as potential predictors of churn. These data elements belonged to two broad categories:

1) Online activities such as how often a subscriber visited the websites and the count of different types of page views. This data could be extracted and aggregated from the datasets received from Google Analytics daily.

2) Subscription history such as tenure days, last payment date and amount, and rate history.

The data engineering team then accessed the feasibility of obtaining these data. As it turned out, the Google Analytics datasets were readily available, but getting the subscription history was very difficult. These were the challenges:

1) With several mergers and acquisitions, subscription transactions were managed in different vendor software applications. These applications had very different data structures, especially for historic data.

2) All the existing data normalizations were by-products of purpose-built data pipelines for very specific financial reporting use cases. And those data pipelines were built over many years.

3) The data engineers who were maintaining those pipelines only knew the code of the pipelines and were not necessarily familiar with the underlying reason for those program logic. In other words, they do not have a good understanding of the data.

4) The data scientists did not know how to link subscriptions with online activities of consumers.

It became evident that using the existing approach to build a data pipeline for this new use case would take many weeks.

Following the Data Exploration workflow in section IIIA, the following steps were taken to make data available for exploration:

1) A data scientist and a data engineer were designated to learn the subscriptions and Google Analytics datasets.

2) Create an ID graph for the Consumer entity. A consumer is identified in Google Analytics by an ID called anonymous ID and is identified by an account number in one of the subscription systems.

3) Using the ID graph, connect Google Analytics and the subscriptions data to the Consumer entity by using anonymous ID and account number as foreign keys.

The resulting people-data-entities knowledge graph, ID graph, and ER-diagram are shown in Figure 8.

When the lightly processed datasets were available, the data scientist and data engineer worked together to write the SQL to obtain the data for experimentation. The SQLs turned out to be long and complex, but it was adequate for data exploration purpose.

Through correlation analysis, the data scientist was able to quickly determine which of the data elements were good churn predictors and use them to develop a prediction model. The data engineer then followed the Data for Concrete Requirements workflow in section IIIB to build a data pipeline to prepare data for the prediction model in production. Much

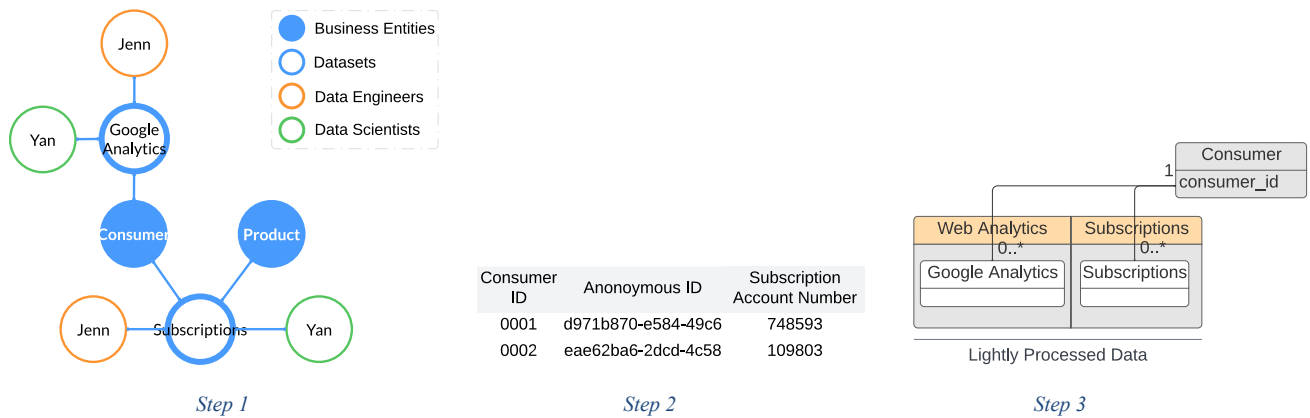

*Step 1*          *Step 2*          *Step 3*

**FIGURE 8.** illustration of the data exploration workflow in the case study.
**Step 1:** The people-data-entities graph after a data scientist and a data engineer were assigned to explore the subscriptions and Google Analytics datasets.
**Step 2:** An ID graph is created for the Consumer entity, linking anonymous IDs from Google Analytics with account numbers from multiple subscription systems.
**Step 3:** Using the ID graph, Google Analytics and subscription datasets are connected to the Consumer entity via foreign keys.

And it was contingent on postponing some of the already planned projects. In essence, it would take many weeks or even months just to prepare data for initial exploration. This was deemed impractical.

*B. THE BSDS DIFFERENCE*

of the SQL code written as well as knowledge gained during data exploration was reusable, and the production pipeline was completed shortly after the data science model was finalized.

Another data pipeline was created to add the model prediction as new profile attributes of the Consumer entity.

*C. RESULT*

The case study was carried out following an iterative 5-step machine learning (ML) engineering process:

1) Frame ML task based on business objective.
2) Data acquisition and preparation.
3) Feature engineering.
4) ML model development.
5) Experimentation and evaluation.

The success metric for the case study was closely aligned with the main value proposal of BSDS: accelerating time to market. To evaluate time to market, the duration required to complete the above five-step ML engineering process was compared against a baseline existing approach. The result is shown in Table 1. It is important to note that this process is inherently iterative. While the total time required to complete all five steps was accurately recorded, there were no precise milestones marking the completion of each individual step. Instead, the durations of the individual steps were measured at the points when those steps were substantially completed.

TABLE I
CASE STUDY METRIC: TIME TO MARKET

| | | Existing Approach | BSDS |
|---|---|---|---|
| 1 | Frame ML task based on business objective | 1 week | 1 week |
| 2 | Data acquisition and preparation | > 6 weeks [a] | 2 weeks |
| 3 | Feature engineering | | 1 week |
| 4 | Develop ML model | | 1 week |
| 5 | Experimentation and evaluation | | 1 week |

End-to-end execution time of BSDS and benchmarked it against an existing workflow.
[a]The original project leveraging the legacy approach was halted at Step 2 due to an estimated data acquisition latency exceeding six weeks, rendering overall project infeasible within the desired timeline.

### D. THEORETICAL FRAMING AND GENERALIZATION

The key principles of BSDS that contributed to the accelerating time to market value proposal validated in this case study are:

- Business semantic centric architecture
- Alignment of datasets with business semantics via the ID graph
- Alignment of data team with business semantics
- A holistic framework to integrate the above three constructs

These principles are transferable across various businesses and industries, provided that certain contextual qualifiers are met. Table 2 describes how each of these principles helps to accelerate time to market and the contextual qualifier necessary for the principle to function across different organizations and industries—achieving the overall generalizability of BSDS.

TABLE 2
TRANSFERABLE PRINCIPLES & CONTEXTUAL QUALIFIERS

| Principle | Why it works | Contextual Qualifier |
|---|---|---|
| Business semantic-centric architecture | Provides shared ontology and points of linkage for datasets | There exist logically identifiable business entities |
| Semantic alignment of datasets via ID graph | Collapses silo boundaries by providing common pre-defined join keys | Stable identifiers exist for business entities |
| Data team alignment with business semantics | Improves collaboration by providing a common view of data that everyone understands | Data team has business acumen skills and data ownership culture |
| A holistic business goal-focus framework | Provides a playbook for efficient planning and execution | Executive buy-in |

### E. DISCUSSION

The case study illustrated that even for a company that possessed a highly skilled data team and a modern data platform leveraging current cloud technologies, it may not be sufficient to adequately support a data driven business. It further revealed that the challenge was not a lack of tools, technologies, or technical expertise. Rather, the core problem was in the existing approach that prioritized platform capabilities and operational expertise over a deeper understanding of the data itself. In contrast, BSDS places business semantics at the forefront. Additionally, the previous approach lacked a framework to unify human, tools and data, whereas BSDS provides a holistic approach that aligns all elements with business semantics to drive successful business outcomes.

## VII. CONCLUSION, LIMITATION AND FUTURE WORK

### A. CONCLUSION

Business Semantics-Centric, AI-Assisted Data System (BSDS) is a comprehensive framework for enabling data-driven businesses. BSDS places business semantics at the core of its architecture, workflows, and team organization and ensures that data availability aligns with business priorities rather than dictating them. Its modular design integrates curated data, a knowledge base, AI agents, and data pipelines. BSDS also highlights the human factor, fostering expertise in key business semantics. This ensures adaptability to evolving needs and effective cross-functional collaboration. Validated in a real-world scenario, BSDS provides a practical, scalable implementation blueprint and offers a foundation for future research. It redefines data system as dynamic enabler of business success, making it essential for modern, data-driven businesses.

### B. LIMITATION

Section V. LOW BARRIER TO ENTRY & SCALABLE highlights how the BSDS core architecture is designed to support scalability as data volume and use case complexity increase. However, the validation case study primarily demonstrates the system's ease of initial adoption, and therefore large-scale scalability and generalizability across business domains remain to be thoroughly tested.

### C. FUTURE WORK

For future research, I recommend these avenues:

1) Building on BSDS's comprehensive system view and its knowledge graph encompassing business entities, datasets, and individuals, researchers can explore data system planning and optimization driven by business priorities. This research can be guided by complex systems and adaptive network theories.

2) Investigating autonomous data systems by utilizing BSDS's well-defined components, including its knowledge base, AI agents, and their interactions.

3) While the current implementation of BSDS focuses on a single organization, its architecture—centered on the ID Graph and semantic data linkage—naturally lends itself to industry-wide adoption. Integrating with blockchain technology could enable secure, decentralized collaboration across multiple companies, with a global ID Graph serving as a shared integration hub. Leveraging a blockchain-based framework, such as [39], would help ensure the integrity and trustworthiness of the global ID Graph, opening the door to new collaborative use cases and cross-organizational collaboration.